\documentclass[letterpaper, 10pt, conference]{ieeeconf}  

\IEEEoverridecommandlockouts                              
\usepackage{times} 
\usepackage{amsmath} 
\usepackage{amssymb}  

\usepackage{booktabs}
\usepackage{graphicx}
\usepackage{multirow}
\usepackage{physics}
\usepackage{subcaption}
\usepackage{xcolor}
\usepackage{bm}
\usepackage{mathtools}
\usepackage{cite}
\usepackage{hyperref}
\hypersetup{hidelinks=true}

\usepackage{amsthm}

\newtheorem{prop}{Proposition}

\theoremstyle{remark}
\newtheorem{defn}{Definition}

\newtheorem{assumption}{Assumption}
\newtheorem{remark}{Remark}

\newcommand{\R}{\mathbb{R}}

\makeatletter
\newcommand\footnoteref[1]{\protected@xdef\@thefnmark{\ref{#1}}\@footnotemark}
\makeatother

\definecolor{fireenginered}{rgb}{0.81, 0.09, 0.13}

\definecolor{amber}{rgb}{1.0, 0.49, 0.0}
\definecolor{Green}{rgb}{0.1, 0.6, 0.0}

\title{\LARGE \bf
    R2DN: Scalable Parameterization of Contracting and\\Lipschitz Recurrent Deep Networks
}

\author{Nicholas H. Barbara, Ruigang Wang, and Ian R. Manchester%
\thanks{*This work was supported in part by the Australian Research Council (DP230101014) and Google LLC.}%
\thanks{The authors are with the Australian Centre for Robotics (ACFR) and the School of Aerospace, Mechanical and Mechatronic Engineering, The University of Sydney, Australia {\tt\small \{nicholas.barbara, ruigang.wang, ian.manchester\}@sydney.edu.au}.}%
}

\begin{document}
\bstctlcite{IEEEexample:BSTcontrol}
\maketitle
\thispagestyle{empty}
\pagestyle{empty}

\begin{abstract}
This paper presents the Robust Recurrent Deep Network (R2DN), a scalable parameterization of stable and robust recurrent neural networks for machine learning and data-driven control. We construct R2DNs as the feedback interconnection of a linear time-invariant system and a 1-Lipschitz deep feedforward network, and directly parameterize the weights so that our models are stable (contracting) and robust to input perturbations (Lipschitz) by design. Our parameterization uses a structure similar to the recurrent equilibrium network (REN), but without having to iteratively solve an equilibrium layer at each time-step. This speeds up model inference and training on GPUs, and makes it computationally feasible to scale up the network size and input sequence length in comparison to RENs. We compare R2DNs to RENs on representative problems in nonlinear system identification, observer design, learning-based feedback control, and sequential image classification. We find that training and inference are up to an order of magnitude faster with similar performance, and that they scale more favorably with respect to model expressivity.
\end{abstract}

\section{Introduction} \label{sec:intro}

Built on the power of deep neural networks (DNNs), modern machine learning has achieved significant progress in a wide range of fields across science and engineering \cite{Jumper++2021,Takahiro++2022}. However, despite their expressive function approximation capabilities, it is widely known that DNNs can be very sensitive to input perturbations, leading to brittle behavior and unexpected failures \cite{Szegedy++2013,Huang++2017,Shi++2024}. This is particularly true for applications involving dynamical systems, where it is natural to consider DNNs with internal states, such as recurrent neural networks (RNNs) \cite{elman1990finding}. In these cases, a fundamental challenge is to regulate not only the input sensitivity, but also the stability of an RNN's internal states over time \cite{Revay++2023}.

\begin{figure}[!t]
    \centering
    \begin{subfigure}[b]{0.49\linewidth}
         \centering
        \includegraphics[trim={5.5cm 18.2cm 11cm 6cm},clip,width=\textwidth]{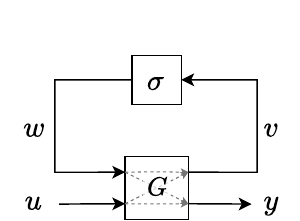}
        \caption{REN architecture \cite{Revay++2023}.}
        \label{fig:arch-ren}
    \end{subfigure}
        \begin{subfigure}[b]{0.49\linewidth}
         \centering
        \includegraphics[trim={11cm 18.2cm 5.5cm 6cm},clip,width=\textwidth]{Images/r2dn_v4.drawio.pdf}
        \caption{R2DN architecture (ours).}
        \label{fig:arch-r2dn}
    \end{subfigure}
    \caption{Block diagrams for the REN and the proposed R2DN architectures. We replace the scalar activation function $\sigma$ with a 1-Lipschitz feedforward network, and modify the LTI system $\bm{G}$ to remove direct feedthrough from $w \rightarrow v$.}
    \label{fig:architectures}
    \vspace{-4mm}
\end{figure}

Several neural architectures have been proposed to address the limitations of RNNs by directly imposing constraints on their internal stability and input-output robustness \cite{Manek+Kolter2019,Revay++2023,kozachkov2022RNNs,jaffe2024learning,massai2025free}. One particular architecture is the \textit{recurrent equilibrium network}, or REN \cite{Revay++2023}, which is structured as the feedback interconnection of a linear time-invariant (LTI) system and a scalar activation function, as shown in Figure~\ref{fig:arch-ren}. The REN model class contains many common network architectures as special cases, including multi-layer perceptrons (MLPs), convolutional neural networks (CNNs), and residual networks (ResNets). RENs are designed to satisfy strong internal stability and input-output robustness properties via contraction \cite{Lohmiller+Slotine1998} and integral quadratic constraints (IQCs) \cite{Megretski+Rantzer1997}, respectively. Most importantly, these stability properties are guaranteed \textit{by construction} via a direct parameterization -- a surjective mapping from a vector of learnable parameters $\theta \in \R^N$ to the network weights and biases -- which makes RENs compatible with standard neural network training algorithms and has enabled their use in a range of tasks including system identification \cite{Revay++2023,shakib2024parameterised}, observer design \cite{zheng2025robust}, stability-certified reinforcement learning \cite{Barbara++2025c,Furieri++2024}, imitation learning \cite{soleimani2025contractive}, and power electronics \cite{feng2025computationally}. 

However, a key limitation of RENs is that calling the model requires iteratively solving an implicit equation for an equilibrium layer \cite{Bai++2019,Revay++2020}, which can be computationally costly both during training and in applications that demand frequent model inference, such as control. Moreover, directly parameterizing RENs with a given sparsity structure (e.g., convolutions) is not straightforward \cite{manchester2026neural}, which further limits the scalability and design flexibility of these models. We seek to address these limitations while retaining the internal stability and input-output robustness properties of RENs.

In this paper, we introduce the \textit{robust recurrent deep network} (R2DN) as a scalable, computationally efficient alternative to RENs \cite{Revay++2023}. Our approach shares a similar structure and direct parameterization to RENs, but with two small tweaks to the model architecture (Figure~\ref{fig:arch-r2dn}) which lead to dramatic improvements in scalability and design flexibility: (a) we eliminate the equilibrium layer by removing the feedthrough term from $w$ to $v$ in the LTI system $\bm{G}$; and (b) we replace the scalar activation $\sigma$ with a scalable 1-Lipschitz DNN. Our numerical experiments show that R2DNs are up to an order of magnitude faster in training and inference than RENs with similar test performance on tasks in system identification, estimation, control, and sequential image classification.

{\bf Related work.} Recent years have seen a boom in the development of scalable RNN architectures, particularly in the large language model (LLM) community, where they have been proposed as a scalable alternative to transformers for long-horizon sequence modeling. In particular, the focus has been on large-scale, structured state-space models (SSMs) \cite{gu2024mamba}, which are linear (often parameter-varying) systems that can be efficiently evaluated on GPUs with parallel associative scans \cite{smith2022Simplified}. 
The LLM community has noted that directly parameterizing these SSMs to be internally stable is critical in preventing model divergence over long sequences \cite{orvieto2023Resurrectinga}. 
Works from the controls community such as \cite{massai2025free,zakwan2026Free} have also proposed parameterizations of stable SSMs which additionally certify $\ell_2$-gain bounds on the input-output response, allowing them to be used as controllers with closed-loop stability guarantees, but they are yet to be tested at scale.

At the same time, there is continued focus on improving the scalability of \textit{nonlinear} RNNs \cite{gonzalez2024Scalable,gonzalez2025Predictability,danieli2025ParaRNN}, since the expressivity of nonlinear interactions permits smaller nonlinear models to solve state-tracking and recall tasks that large linear models cannot \cite{danieli2025ParaRNN}. 
In the nonlinear world, scalable parameterizations of stable RNNs have received comparatively little attention, with notable exceptions including \cite{kozachkov2022RNNs,farsang2025Parallelization}.
The present paper addresses this gap by focusing on scalable parameterizations of not only stable, but also robust nonlinear RNNs. We choose the incremental notions of contraction for internal stability \cite{Lohmiller+Slotine1998} and Lipschitzness (incremental $\ell_2$-gain) for input-output robustness. These incremental properties ensure both smooth and (under mild assumptions \cite{Lohmiller+Slotine1998}) bounded responses to perturbations, facilitating smooth generalization to unseen data \cite{manchester2026neural}.

\section{Problem Setup} \label{sec:prob}
Given a dataset $\mathcal{D}$, we consider the problem of learning a nonlinear state-space model of the form
\begin{equation}\label{eqn:model}
    x_{t+1}=f_\theta(x_t,u_t),\quad y_t=h_\theta(x_t, u_t),
\end{equation}
where $x_t\in \R^{n}, u_t\in \R^m, y_t\in \R^p$ are the states, inputs, and outputs of the system at time $t\in \mathbb{N}$, respectively. Here $f_\theta: \R^n\times \R^m\rightarrow\R^n$ and $h_\theta: \R^n\times \R^m\rightarrow\R^p$ are parameterized by some learnable parameter vector $\theta \in \Theta\subseteq \R^N$ (e.g., the weights and biases of a deep neural network). The learning problem can be formulated as an optimization problem
\begin{equation}\label{eqn:learning}
    \min_{\theta \in \Theta}\quad \mathcal{L}(f_\theta, h_\theta; \mathcal{D})
\end{equation}
for some loss function $\mathcal{L}$.

If $\Theta = \mathbb{R}^N$, the learning problem \eqref{eqn:learning} has no constraints and can be solved via unconstrained optimization tools such as gradient descent, which is particularly useful when working with large models. If, however, $\Theta \subset \mathbb{R}^N$, the model must be trained with constrained solvers, which are typically computationally expensive and limit scalability \cite{manchester2026neural}.
The focus of this paper is therefore to directly parameterize stable and robust nonlinear models \eqref{eqn:model} in the following sense.
\begin{defn}
    A model parameterization $\mathcal{M}:\theta\mapsto (f_\theta, h_\theta)$ with $\theta \in \Theta$ is called a \emph{direct parameterization} if $\Theta=\R^N$.
\end{defn}

We use contraction \cite{Lohmiller+Slotine1998} as our notion of internal stability. Contraction is a stability definition that is independent of any particular equilibrium point or trajectory, which is ideal for models required to generalize to unseen initial states and operating conditions. One can think of a contracting system as exponentially forgetting its initial states as follows.
\begin{defn} \label{dfn:contraction}
    A model \eqref{eqn:model} is said to be \emph{contracting} with rate $\alpha\in [0,1)$ and overshoot $K>0$ if for any two initial states $ x_0^a, x_0^b\in\R^n$ and the same input sequence $u\in \ell^m$, the state sequences $ x^a $ and $ x^b $ satisfy 
    \begin{equation}\label{eqn:contraction} 
        |x_t^a-x_t^b|\leq K\alpha^t|x_0^a - x_0^b|,\quad \forall t\in \mathbb{N}
    \end{equation}
    where $|(\cdot)|$ denotes the Euclidean norm, and $\ell^m$ denotes the set of sequences $u: \mathbb{N} \rightarrow \mathbb{R}^m$.
\end{defn}
Beyond internal stability, we quantify the input-output robustness of models \eqref{eqn:model} via IQCs and Lipschitz bounds.

\begin{defn} \label{dfn:iqc}
    A model \eqref{eqn:model} is said to admit the \emph{incremental integral quadratic constraint} (incremental IQC) defined by $(Q,S,R)$ with $Q=Q^\top\in \R^{p\times p}$, $S\in \R^{m\times p}$, $R=R^\top\in \R^{m\times m}$, if for all pairs of solutions with  initial conditions $ x_0^a, x_0^b\in\R^n $ and input sequences $ u^a,u^b\in \ell^m $, the output sequences $ y^a ,y^b\in \ell^p$ satisfy
	\begin{equation}\label{eqn:iqc}
		\sum_{t=0}^{T}\begin{bmatrix}
		y_t^a - y_t^b \\ u_t^a-u_t^b
		\end{bmatrix}^\top
		\begin{bmatrix}
		Q & S^\top \\ S & R
		\end{bmatrix}
		\begin{bmatrix}
		y_t^a - y_t^b \\ u_t^a-u_t^b
		\end{bmatrix}\geq -d(x_0^a, x_0^b),
	\end{equation} 
	$\forall T \in \mathbb{N}$, and for $d(x_0^a, x_0^b) \geq 0$ with $d(x_0^a, x_0^a) = 0$.
\end{defn} 
A model \eqref{eqn:model} is said to be \emph{$\gamma$-Lipschitz} with $\gamma>0$ if it satisfies the incremental IQC defined by $(-\frac{1}{\gamma} I,0, \gamma I)$. The Lipschitz bound $\gamma$ (incremental $\ell_2$-gain bound) then quantifies a model's output sensitivity to input perturbations.

\section{Review of Recurrent Equilibrium Networks}\label{sec:ren}

RENs \cite{Revay++2023} are written as a Lur'e system (Figure~\ref{fig:arch-ren}). They are a feedback interconnection of a learnable LTI system $\bm{G}$ and a fixed scalar activation $\sigma$ with its slope restricted in $[0,1]$ (e.g., ReLU or tanh) acting element-wise on vectors:
\begin{subequations} \label{eqn:ren}
    \begin{align}
        \begin{bmatrix}
            x_{t+1} \\ v_t \\ y_t
        \end{bmatrix}&=
        \overset{W}{\overbrace{
        \left[
            \begin{array}{c|cc}
            A & B_1 & B_2 \\ \hline 
            C_{1} & D_{11} & D_{12} \\
            C_{2} & D_{21} & D_{22}
        \end{array} 
        \right]
        }}
        \begin{bmatrix}
            x_t \\ w_t \\ u_t
        \end{bmatrix}+
        \overset{b}{\overbrace{
            \begin{bmatrix}
                b_x \\ b_v \\ b_y
            \end{bmatrix}
        }} \label{eqn:ren-lti}\\
        w_t&=\sigma(v_t), \label{eqn:ren-activation}
    \end{align}
\end{subequations}
where $v_t, w_t \in \mathbb{R}^q$ are the neuron input and output variables, and $(W, b)$ are learnable weights and biases. The feedthrough term $D_{11}$ forms an equilibrium layer (implicit layer) as
\begin{equation}\label{eqn:implicit-layer}
    w_t=\sigma(D_{11}w_t+C_1x_t +D_{12}u_t+b_v).
\end{equation}
With $D_{11}=0$, the REN \eqref{eqn:ren} is simply a single-layer network. For nonzero $D_{11}$, however, the equilibrium layer \eqref{eqn:implicit-layer} contains a rich set of multi-layer feedforward network architectures as special cases -- see \cite{ghaoui2019implicit,Revay++2020}, and \eqref{eqn:r2dn-as-ren} below for examples.

If the equilibrium layer \eqref{eqn:implicit-layer} is well-posed (i.e., for any $(x_t,u_t)$ there exists a unique solution $w_t$) as addressed in \cite{Revay++2023}, then it represents a static map $\phi_\mathrm{eq}: (x,u)\mapsto w$ and the REN \eqref{eqn:ren} can be written in the form \eqref{eqn:model} with $f_\theta, h_\theta$ given by
\begin{equation}\label{eqn:fh-ren}
    \begin{split}
        f_\theta(x,u) &= Ax  + B_1 \phi_\mathrm{eq}(x, u) + B_2 u + b_x,\\
        h_\theta(x,u) &= C_2 x + D_{21} \phi_\mathrm{eq}(x,u) + D_{22} u  + b_y.
    \end{split}
\end{equation}
A central result of \cite{Revay++2023} was the direct parameterization of RENs that are both contracting and  satisfy general $(Q,S,R)$-type incremental IQCs in the sense of Definitions~\ref{dfn:contraction} and \ref{dfn:iqc}.

\section{Robust Recurrent Deep Network (R2DN)}\label{sec:r2dn}

Our proposed R2DNs have the same structure as RENs but for the two key differences illustrated in Figure~\ref{fig:architectures}: we remove the equilibrium layer \eqref{eqn:implicit-layer} by setting $D_{11} = 0$; and we allow the static nonlinearity to be any DNN $\phi_g$ rather than just a scalar activation $\sigma$.
These design choices are not merely engineering tricks to eliminate the equilibrium layer, but rather major architectural changes  allowing us to scale up the model while maintaining the same stability and robustness guarantees available to RENs.

The R2DN model structure is defined as
\begin{subequations} \label{eqn:r2dn}
    \begin{align}
        \begin{bmatrix}
            x_{t+1} \\ v_t \\ y_t
        \end{bmatrix} &=
        \overset{W}{\overbrace{
		\left[
            \begin{array}{c|cc}
            A & B_1 & B_2 \\ \hline 
            C_{1} & 0 & D_{12} \\
            C_{2} & D_{21} & D_{22}
		\end{array} 
		\right]
        }}
        \begin{bmatrix}
            x_t \\ w_t \\ u_t
        \end{bmatrix}+
        \overset{b}{\overbrace{
            \begin{bmatrix}
                b_x \\ b_v \\ b_y
            \end{bmatrix}
        }} \label{eqn:r2dn-lti}\\
        w_t &=\phi_g(v_t), \label{eqn:r2dn-layer}
    \end{align}
\end{subequations}
where the learnable parameters are $(W,b,g)$ with $g\in \R^{M}$ as the free parameters of the static DNN $\phi_g$. This system can be rewritten in the form \eqref{eqn:model} with
\begin{equation}\label{eqn:fh}
    \begin{split}
        f_\theta(x,u) &= Ax  + B_1 \phi(x,u) + B_2 u + b_x,\\
    h_\theta(x,u) &= C_2 x + D_{21} \phi(x,u) + D_{22} u  + b_y,
    \end{split}
\end{equation}
where $\phi(x,u)=\phi_g(C_1 x +D_{12}u + b_v)$. 

If the static network $\phi_g$ can be represented by an equilibrium network \eqref{eqn:implicit-layer}, the R2DN can be reformulated exactly as a sparsely structured REN with \textit{nonzero} $D_{11}$. We illustrate this below for the case where $\phi_g$ is an $L$-layer feedforward network (such as an MLP or CNN) with the form
\begin{equation*}
z_0=v_t,\;
z_{l+1}=\sigma(W_l z_l+b_l),\;
\phi_g(v_t)=W_Lz_L+b_L,
\end{equation*}
where $z_l$ is the $l^\text{th}$ hidden unit and $l=0,\ldots,L-1$. Let $W, b$ denote the learnable LTI parameters of the original R2DN, and let $\bar{W}, \bar{b}$ be those of its equivalent REN. Then the weights $A, B_2, C_2, D_{22}$ are unchanged and 
\begin{gather}
\bar{B}_1 = \begin{bmatrix} 0 & \cdots & 0 & B_1W_L \end{bmatrix}, \quad
\bar{b}_x = b_x + B_1 b_L, \notag \\
\bar{D}_{21} = \begin{bmatrix} 0 & \cdots & 0 & D_{21}W_L \end{bmatrix}, \quad
\bar{b}_y = b_y + D_{21} b_L, \notag \\
\bar{C}_1 = \begin{bmatrix} W_0 C_1 \\ 0 \\ \vdots \\ 0 \end{bmatrix}, \quad
\bar{D}_{12} = \begin{bmatrix} W_0 D_{12} \\ 0 \\ \vdots \\ 0 \end{bmatrix},
\label{eqn:r2dn-as-ren} \\
\bar{D}_{11} =
    \begin{bmatrix}
        0 & & & \\
        W_1 & \ddots & & \\
        \vdots & \ddots & 0 & \\
        0 & \cdots & W_{L-1} & 0
    \end{bmatrix}, \quad
\bar{b}_v = \begin{bmatrix} W_0 b_v + b_0 \\ b_1 \\ \vdots \\ b_{L-1} \end{bmatrix}. \notag
\end{gather}
The key advantage of the R2DN model structure \eqref{eqn:r2dn} is that we can directly parameterize it to be contracting and Lipschitz (see Section~\ref{sec:direct-param} for details), whereas it is not obvious how to do so for a REN with the particular structure in \eqref{eqn:r2dn-as-ren}. 
Moreover, the R2DN architecture permits the use of nonlinearities $\phi_g$ that cannot be represented by an equilibrium layer \eqref{eqn:implicit-layer}, such as transformers \cite{vaswani2017attention}, whose attention layers have no equilibrium-layer representation.

We now make the following assumption about $\phi_g$, leading to the result in Proposition~\ref{prop:1}, which will allow us to directly parameterize stable and robust R2DNs in Section~\ref{sec:direct-param}.
\begin{assumption}\label{asmp:phi}
    $\phi_g:\R^q\rightarrow\R^q$ is 1-Lipschitz for any $g\in \R^{M}$.
\end{assumption}

\begin{prop}\label{prop:1}
    Suppose that Assumption~\ref{asmp:phi} holds, and \eqref{eqn:r2dn-lti} is contracting and admits the incremental IQC defined by 
    \begin{equation} \label{eqn:qsr-small-gain}
        \tilde{Q} = \mqty[-I & 0 \\ 0 & Q], \ \ \tilde{S} = \mqty[0 & 0 \\ 0 & S], \ \ \tilde{R} = \mqty[I & 0 \\ 0 & R]
    \end{equation}
    with $0 \succ Q \in \R^{p\times p}$, $S\in \R^{m\times p}$ and $R=R^\top \in \R^{m\times m}$. Then, \eqref{eqn:r2dn} is contracting and admits the incremental IQC defined by $(Q,S,R)$.
\end{prop}

\begin{proof}  
The proof is provided in \cite[Sec.~6.3]{barbara2026thesis} and follows from the incremental small-gain theorem and standard dissipativity arguments for incremental IQCs \cite{Megretski+Rantzer1997,manchester2026neural}.
\end{proof}

\begin{remark}
Contraction is achieved in Proposition~\ref{prop:1} with a version of the small-gain theorem requiring $\phi_g$ to be 1-Lipschitz, and the LTI system $\bm{G}$ in \eqref{eqn:r2dn-lti} to have a Lipschitz bound less than one from $w \mapsto v$. This stability condition is not as conservative as it may seem because we jointly learn $\bm{G},\phi_g$.
To illustrate, suppose (with $y_t, u_t$ dimension zero) we have an interconnection of $\bm{G},\phi_g$ which admits a $(Q,S,R)$-type IQC but does not satisfy \eqref{eqn:qsr-small-gain} or Assumption~\ref{asmp:phi}. A standard approach \cite{Megretski+Rantzer1997} to reduce conservatism on the stability of the interconnection is to introduce invertible multiplier matrices $M, N\in\R^{q\times q}$ such that for any $a,b\in\R^q$,
$$
\|M\bm{G}N^{-1}\|_\infty <1, \ |N\phi(M^{-1}a) - N\phi(M^{-1}b)| \le |a-b|,
$$
since the $\mathcal{H}_\infty$ norm of an LTI system is its Lipschitz bound.
Then, the interconnection of $M\bm{G}N^{-1}$ and $N\phi_g(M^{-1}x)$ is stable by Proposition~\ref{prop:1}. That is, $M,N$ are effectively absorbed into the learned representations of $\bm{G}$ and $\phi_g$.
\end{remark}

\begin{remark}
Assumption~\ref{asmp:phi} implies that $\phi_g$ satisfies the incremental IQC defined by $(-I, 0, I)$. This can be further extended to other $(Q,S,R)$-type IQCs, such as incrementally passive or strongly monotone neural networks \cite{Wang++2024b}.
\end{remark}

The main implication of Proposition~\ref{prop:1} is that the R2DN parameterization can be divided into two separate parts: 
\begin{itemize}
    \item[(a)] parameterizations of 1-Lipschitz DNNs $\phi_g$; and
    \item[(b)] parameterizations of the LTI system $\bm{G}$ in \eqref{eqn:r2dn-lti} subject to the IQC defined by \eqref{eqn:qsr-small-gain}.
\end{itemize}
Many direct parameterizations of 1-Lipschitz neural networks already exist for a variety of DNN architectures \cite{miyato2018spectral, trockman2021orthogonalizing, Wang+Manchester2023, Pauli++2024,araujo2023unified,qi2023lipsformer}. We therefore leverage the existing Lipschitz DNNs for part (a) and focus on part (b) for the remainder of this paper.

\section{Direct Parameterization of R2DN} \label{sec:direct-param}
We now present direct parameterizations of contracting and Lipschitz R2DNs by constructing LTI systems \eqref{eqn:r2dn-lti} based on \eqref{eqn:qsr-small-gain}. Our parameterizations are closely related to those of RENs in \cite[Sec.~V]{Revay++2023} but with new insight for when $D_{11} = 0$.

\subsection{Robust LTI Systems}
\vspace{-0.5mm}

Since \eqref{eqn:r2dn-lti} is LTI, we first introduce sufficient conditions for LTI systems satisfying the assumptions in Proposition~\ref{prop:1}. We seek contracting LTI systems $\bar{\bm{G}}:\bar{u}\mapsto \bar{y}$ admitting the incremental IQC defined by $(\bar{Q},\bar{S},\bar{R})$ with
\begin{equation}\label{eqn:QSR}
    \bar{Q} \prec 0,\quad \bar{R} - \bar{S} \bar{Q}^{-1} \bar{S}^\top \succ 0.
\end{equation}
We consider state-space realizations
\begin{equation} \label{eqn:lti}
    \mqty[x_{t+1} \\ \bar{y}_t] = \mqty[A & B \\ C & D] \mqty[x_t \\ \bar{u}_t]
\end{equation}
and over-parameterize the system as follows
\begin{equation} \label{eqn:lti-implicit}
    \mqty[Ex_{t+1} \\ \bar{y}_t] = \mqty[\mathcal{A} & \mathcal{B} \\ C & D] \mqty[x_t \\ \bar{u}_t],
\end{equation}
where $\mathcal{A} = E A$ and $\mathcal{B} = E B$ with $E$ an invertible matrix. Then, we have the following result.

\begin{prop} \label{prop:lti}
    Suppose that $(\bar{Q},\bar{S},\bar{R})$ satisfy \eqref{eqn:QSR}. Then, \eqref{eqn:lti} is contracting and admits the incremental IQC defined by $(\bar{Q}, \bar{S}, \bar{R})$ if there exists a $\mathcal{P} \succ 0$ and an invertible $E$ s.t.
    \begin{gather}
        H  \succ 
        \mqty[\mathcal{C}^\top \\ \mathcal{B}] \mathcal{R}^{-1}
        \mqty[\mathcal{C}^\top \\ \mathcal{B}]^\top 
         - 
        \mqty[C^\top \\ 0] \bar{Q} \mqty[C^\top \\ 0]^\top, \label{eqn:H-inq} \\
        \mathcal{R}:= \bar{R}+\bar{S}D + D^\top \bar{S}^\top + D^\top \bar{Q} D \succ 0, \label{eqn:R-inq}
    \end{gather}
    where $ \mathcal{C} = (D^\top \bar{Q}+\bar{S})C$ and 
    \begin{equation}\label{eqn:H-mat}
    H:=
    \begin{bmatrix}
       E^\top + E - \mathcal{P} & \mathcal{A}^\top \\  \mathcal{A} & \mathcal{P}
    \end{bmatrix}.
\end{equation}
\end{prop}

\begin{proof}
    The result follows as a special case of \cite[Thm.~3]{Revay++2023} for RENs with $B_1, C_1, D_{11}, D_{12}, D_{21}, b_v = 0$ in \eqref{eqn:ren}. 
\end{proof}

\subsection{Direct Parameterization of Contracting R2DNs} \label{sec:direct-contraction}
\vspace{-0.5mm}

We now use this result to directly parameterize \eqref{eqn:r2dn-lti} such that its corresponding R2DN \eqref{eqn:r2dn} is contracting. Consider two trajectories of \eqref{eqn:r2dn-lti} with the same input $u$ and different initial states $x_0^a,x_0^b$. The difference dynamics are
\begin{equation}\label{eqn:lti-contraction}
    \begin{bmatrix}
        \Delta x_{t+1} \\ \Delta v_t
    \end{bmatrix}=
    \begin{bmatrix}
        A & B_1 \\ C_1 & 0
    \end{bmatrix}
    \begin{bmatrix}
        \Delta x_t \\ \Delta w_t
    \end{bmatrix}.
\end{equation}
If the difference dynamics admit the incremental IQC defined by $(-I, 0, I)$, then R2DN \eqref{eqn:r2dn} is contracting by Proposition~\ref{prop:1}. 

By comparing the form of \eqref{eqn:lti-contraction} with \eqref{eqn:lti} we have $D=0$ and hence $\mathcal{R}=I$, $\mathcal{C}=0$. Condition~\eqref{eqn:H-inq} then becomes
\begin{equation}\label{eqn:H-contract}
    H\succ \begin{bmatrix}
        C_1^\top C_1 & 0 \\
        0 & \mathcal{B}_1 \mathcal{B}_1^\top 
    \end{bmatrix}
\end{equation}
where $\mathcal{B}_1= E B_1$. We introduce a set of free variables
\[
\{ X\in \R^{2n\times 2n}, \,Y\in \R^{n\times n}, \,\mathcal{B}_1 \in \R^{n\times q},\, C_1 \in \R^{q\times n}\}
\]
then construct and partition $H$ as follows,
\begin{equation} \label{eqn:contraction-param}
        H = X^\top X + \epsilon I +
        \mqty[C_1^\top C_1 & 0 \\ 0 & \mathcal{B}_1 \mathcal{B}_1^\top]=: \mqty[H_{11} & H_{21}^\top \\ H_{21} & H_{22}] ,
\end{equation}
which satisfies \eqref{eqn:H-contract} by design, where $H_{11}, H_{22}\in \R^{n\times n}$ and $\epsilon>0$ is a small constant. We then construct $A, B_1$ from $H$:
\begin{equation}\label{eqn:EAB}
    \begin{split}
        E &= \frac{1}{2}(H_{11} + H_{22} + Y - Y^\top), \\
        A &= E^{-1} H_{21}, \quad B_1 = E^{-1} \mathcal{B}_1,
    \end{split}
\end{equation}
where $E$ is invertible for all $Y$.  
The remaining parameters $\{B_2, D_{12}, C_2, D_{21},D_{22},b\}$ in \eqref{eqn:r2dn-lti} are free to learn as they do not affect the contraction condition \eqref{eqn:H-contract}. Thus, the parameterization satisfies Proposition~\ref{prop:lti} and the resulting R2DN \eqref{eqn:r2dn} is contracting by construction. Note that this is a special case of the robust REN parameterization \cite{Revay++2023} without nonlinearity.

\subsection{Direct Parameterization of $\gamma$-Lipschitz R2DNs} \label{sec:direct-lipschitz}

To also impose bounds on the model's input-output robustness, we can directly parameterize \eqref{eqn:r2dn-lti} such that its R2DN is both contracting and $\gamma$-Lipschitz, where $\gamma>0$ is a hyperparameter to be chosen or learned.
Similar to the previous section, we first consider two trajectories of \eqref{eqn:r2dn-lti} with the input pair $u^a, u^b$ and different initial states $x_0^a,x_0^b$. Their difference dynamics satisfy the LTI system \eqref{eqn:lti} with
\begin{equation} \label{eqn:abcd-lipschitz}
    B=\begin{bmatrix}
        B_1 & B_2
    \end{bmatrix},\; C=\begin{bmatrix}
        C_1 \\ C_2
    \end{bmatrix},\; D=\begin{bmatrix}
        0 & D_{12} \\ D_{21} & D_{22}
    \end{bmatrix}.
\end{equation}
By Proposition~\ref{prop:1}, the R2DN \eqref{eqn:r2dn} is contracting and $\gamma$-Lipschitz if \eqref{eqn:lti} admits the incremental IQC defined by
$$
\bar{Q}=\begin{bmatrix}
    -I & 0\\
    0 & -\frac{1}{\gamma} I
\end{bmatrix}, \quad \bar{S}=0, \quad \bar{R}=\begin{bmatrix}
    I & 0 \\
    0 & \gamma I
\end{bmatrix}.
$$
If $D$ in \eqref{eqn:abcd-lipschitz} were dense, we could take the robust REN parameterization directly from \cite{Revay++2023}. Instead, we must ensure that the upper-left block of $D$ in \eqref{eqn:abcd-lipschitz} is zero (i.e., $D_{11} = 0$). 
From Proposition~\ref{prop:lti} we therefore require
\begin{equation} \label{eqn:R-lmi}
        \mathcal{R} = 
        \begin{bmatrix}
        I -\frac{1}{\gamma} D_{21}^\top D_{21} & -\frac{1}{\gamma} D_{21}^\top D_{22}\\
        -\frac{1}{\gamma} D_{22}^\top D_{21} & \gamma I - \frac{1}{\gamma} D_{22}^\top D_{22} - D_{12}^\top D_{12}
    \end{bmatrix}\succ 0.
\end{equation}
Since it is not trivial to directly parameterize all $D_{12}, D_{21} $ and $D_{22}$ satisfying the above condition, we instead give a direct parameterization of a subset with $D_{22}=0$. In this special case, \eqref{eqn:R-lmi} reduces to
\begin{equation} \label{eqn:R-lipschitz-D22zero}
    \mathcal{R} = \mqty[I - \frac{1}{\gamma}D_{21}^\top D_{21} & 0 \\ 0 & \gamma I - D_{12}^\top D_{12}] \succ 0,
\end{equation}
which is equivalent to $\mathcal{D}_{12}^\top \mathcal{D}_{12} \prec I$ and $\mathcal{D}_{21}^\top \mathcal{D}_{21} \prec I$ with 
\[
\mathcal{D}_{12}=\frac{1}{\sqrt{\gamma}}D_{12},\quad \mathcal{D}_{21}=\frac{1}{\sqrt{\gamma}}D_{21}.
\]
Suitable parameterizations of $\mathcal{D}_{12}$ and $\mathcal{D}_{21}$ can be obtained via the Cayley transformation (see \cite{Wang+Manchester2023} for details). The remaining condition \eqref{eqn:H-inq} in Proposition~\ref{prop:lti} becomes
\begin{equation}\label{eqn:H-lipschitz}
    H\succ \Gamma \mathcal{R}^{-1} \Gamma^\top+ \mqty[C_1^\top C_1 + \frac{1}{\gamma} C_2^\top C_2 & 0\\ 0 & 0]
\end{equation}
with
\begin{equation}
    \Gamma = \mqty[-\frac{1}{\gamma} C_2^\top D_{21} & - C_1^\top D_{12} \\ \mathcal{B}_1 & \mathcal{B}_2].
\end{equation}
The remaining steps follow a similar procedure to Sec.~\ref{sec:direct-contraction}. 

\begin{remark}
Even when $D_{22}=0$, R2DN can still incorporate a direct feedthrough from $u$ to $y$ via the static DNN $\phi_g$ (e.g., when $\phi_g$ is a Lipschitz residual network). However, enforcing $D_{22}=0$ may limit the model expressivity when a tight Lipschitz bound is imposed. We leave a full parameterization of Lipschitz R2DNs with $D_{22}\ne 0$ to future work.
\end{remark}

%
%
\section{Qualitative Comparison of RENs and R2DNs}\label{sec:discussion}

The direct parameterization of RENs in \cite{Revay++2023} and R2DNs in Section~\ref{sec:direct-param} ensure that the resulting models are contracting and Lipschitz by construction. The key architectural decision that separates the two is setting $D_{11} = 0$ for R2DNs. We summarize the advantages of this choice below.

\paragraph{Efficient GPU computation} 
For RENs, solving \eqref{eqn:implicit-layer} with general $D_{11}$ is slow and involves iterative equilibrium solvers (see \cite{Revay++2020}) which can be computationally prohibitive for large-scale models. If $D_{11}$ is parameterized to be strictly lower-triangular as in \cite{Revay++2023}, then \eqref{eqn:implicit-layer} can be solved row-by-row (see Appendix~\ref{app:acyclic-solver}), which provides a significant speed boost on CPUs. However, such a solver remains inefficient on GPUs, which are designed to leverage massive parallelism rather than sequential computation. R2DNs do not have to solve an equilibrium layer and can take full advantage of modern GPU architectures for efficient computation.

\paragraph{Design flexibility}
The proposed parameterization is flexible in that we can choose $\phi_g$ to be any 1-Lipschitz feedforward network. This opens up the possibility of using network structures such as MLPs \cite{miyato2018spectral,Wang+Manchester2023}, CNNs \cite{trockman2021orthogonalizing,Pauli++2024}, ResNets \cite{araujo2023unified}, or transformer-like architectures \cite{qi2023lipsformer}. In contrast, the REN parameterization in \cite{Revay++2023} only allows for $D_{11}$ with particular structures (full or strictly lower-triangular). While, in principle, the REN \eqref{eqn:ren} contains many of the above network architectures, it is not obvious how to parameterize a well-posed contracting and Lipschitz REN with a structured equilibrium layer (e.g., convolution operator). This limits the application of RENs in high-dimensional problems involving audio or visual data, while R2DNs have no such restriction.

\paragraph{Model scalability}
RENs typically have many more parameters than R2DNs given the same number of neurons due to the structure of the equilibrium layer. For a REN with $k$ neurons, the number of parameters in $\phi_\mathrm{eq}$ is proportional to $k^2$. However, for an R2DN with an $L$-layer, $k$-neuron MLP, the number of parameters in $\phi_g$ is proportional to $k^2/L$. The nonlinear block in an R2DN therefore has $\sqrt{L}$ times as many neurons as a comparably sized equilibrium layer.

\begin{remark}
    For problems requiring models with large state dimensions $n$, it may also be desirable for the LTI component \eqref{eqn:r2dn-lti} to have a highly scalable parameterization in addition to the nonlinear component $\phi_g$. The number of learnable parameters scales proportionally to $n^2$ in the parameterizations of both RENs and R2DNs due to the $X^\top X$ terms in \eqref{eqn:contraction-param}. There are several options to mitigate this, two of which are:
    \begin{enumerate}
        \item \textbf{Low-rank parameterization:} Introduce new parameters $\delta \in \R^{2n}$ and $\bar{X}\in \R^{\nu \times 2n}$ with $\nu \ll n$, then replace $X^\top X$ with $\bar{X}^\top \bar{X} + \mathrm{diag}(|\delta|)$ in \eqref{eqn:contraction-param}. The LTI system \eqref{eqn:r2dn-lti} remains dense, but the number of learnable parameters scales linearly with $n$.
        \item \textbf{Parallel components:} Replace \eqref{eqn:r2dn-lti} with many smaller, parallel LTI systems which can be separately or jointly parameterized. In the limit that each system is scalar, the number of parameters scales linearly with $n$. Parallel interconnections of 1-Lipschitz systems remain 1-Lipschitz, so our parameterization remains valid.
    \end{enumerate}
    We leave a detailed study of the effect of scalable LTI parameterizations in RENs and R2DNs to future work.
\end{remark}

%
%
\section{Numerical Experiments} \label{sec:experiment}

We now study the computational benefits of R2DNs over RENs in numerical experiments. All experiments\footnote{\label{note1}\url{https://github.com/nic-barbara/R2DN}} were performed in Python/JAX on an NVIDIA GeForce RTX 5090 GPU. R2DN models were implemented with 1-Lipschitz MLPs constructed from Sandwich layers \cite{Wang+Manchester2023}, and RENs were implemented with strictly lower-triangular $D_{11}$ as outlined in Appendix~\ref{app:acyclic-solver}. We focus our study on contracting RENs and R2DNs in this preliminary investigation, saving a broader comparative study including Lipschitz-bounded models and other model architectures for future work.

\subsection{Scalability and Expressive Power} \label{sec:exp-scalability}

\begin{figure}[!t]
    \centering
    \includegraphics[trim={0cm 0.3cm 0cm 0cm},clip,width=0.9\linewidth]{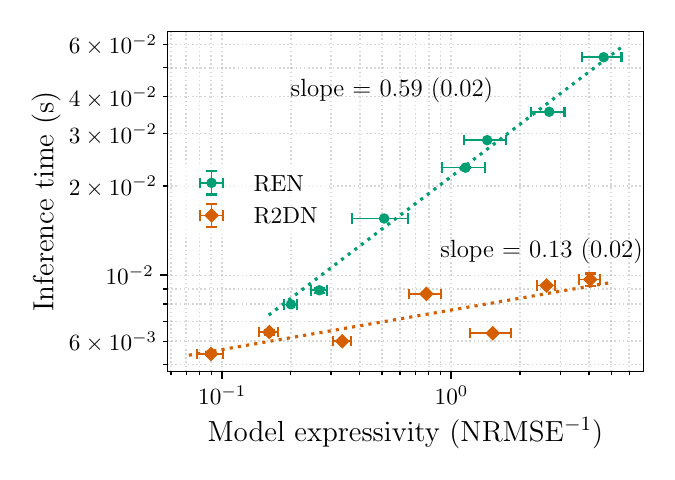}
    \caption{Inference time vs model expressivity. Error bars show one standard deviation across 5 random seeds. Slope standard deviations are in parentheses. Backpropagation scales similarly (REN slope $0.47\pm 0.02$, R2DN slope $0.085\pm0.003$, figure omitted to save space).}
    \label{fig:scaling-relation}
\end{figure}

We first show that computation time for R2DNs scales more favorably with respect to the network's expressive power (expressivity) compared to RENs. We use the heuristic for network expressivity outlined below -- using the total number of learnable parameters or activations would be inappropriate since they could be distributed in many different ways between the components of similarly sized networks.

\begin{figure*}[!t]
    \centering
    \begin{subfigure}[b]{0.32\linewidth}
        \centering
        \includegraphics[trim={0cm 0.5cm 0cm 0cm},clip,width=\textwidth]{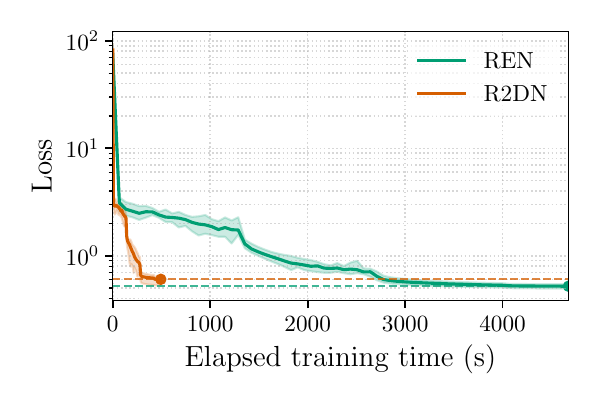}
        \caption{System identification.}
        \label{fig:loss-sysid}
    \end{subfigure}
    \begin{subfigure}[b]{0.32\linewidth}
        \centering
        \includegraphics[trim={0cm 0.5cm 0cm 0cm},clip,width=\textwidth]{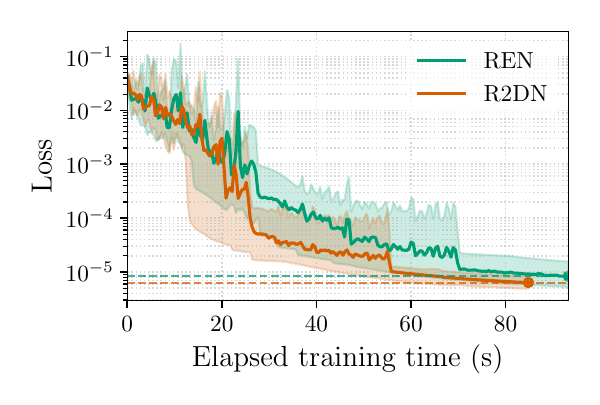}
        \caption{PDE observer design.}
        \label{fig:loss-observer}
    \end{subfigure}
    \begin{subfigure}[b]{0.32\linewidth}
        \centering
        \includegraphics[width=0.92\textwidth]{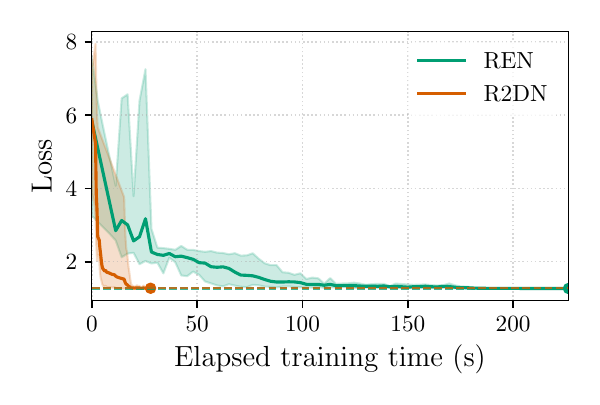}
        \caption{Learning-based feedback control.}
        \label{fig:loss-youla}
    \end{subfigure}
    \caption{Mean loss curves as a function of training wall time for the three benchmark problems in \cite{Revay++2023}. Bands show the loss range over 10 random seeds. Dashed lines show the final mean loss achieved by each model class. }
    \label{fig:performance}
    \vspace{-2mm}
\end{figure*}

We fit the internal dynamics $f_\theta$ from \eqref{eqn:fh-ren} for RENs and \eqref{eqn:fh} for R2DNs to a scalar nonlinear function
\begin{multline*}
f(x,u) = 0.05x + 0.2\sin(x) + u + 0.05\cos(2x_u) + \\ 0.05\sin(3x_u) + 
            0.075 \sin(4x_u) \tan^{-1}(0.1x_u^2)
\end{multline*}
where $x_u := x+u$. The function has a maximum slope w.r.t. $x$ of $|\pdv{}{x}f(x,u)| \lessapprox0.9$. It is not in either of the REN or R2DN model classes, but can be approximated given a sufficiently large number of neurons in $\phi_\mathrm{eq}$ or $\phi_g$, respectively. We then computed the normalized root-mean-square test error
$$
\text{NRMSE} = \frac{| f(x,u) - f_\theta(x, u) |}{|f(x,u)|} \times 100
$$
for test batches of $x,u$ and took $1 / \text{NRMSE}$ as a measure of the network's expressive power. Further experiment details are provided in Appendix~\ref{app:expressivity-details}.

The results in Figure~\ref{fig:scaling-relation} show how mean computation time scales with model expressivity for each network architecture. Computation time was measured by evaluating the mean inference and backpropagation (gradient calculation) time over 1000 function calls for each model, using sequences of length 128 with a batch size of 64. In both cases, computation time increases with model expressivity. However, the increase occurs at a much faster rate for the RENs, whereas R2DNs can clearly scale to more expressive models with minimal increase in training and inference time. This bodes well for future applications of R2DNs to large-scale problems which require much larger recurrent models.

\subsection{Training Speed and Test Performance} \label{sec:exp-performance}

We now compare the performance of each model class on the three case studies introduced in \cite{Revay++2023} for RENs, in addition to sequence modeling tasks for image classification:
\begin{enumerate}
    \item Stable and robust nonlinear system identification on the F-16 ground vibration dataset \cite{noel2017f16};
    \item Learning nonlinear observers for a reaction-diffusion partial differential equation (PDE);
    \item Data-driven, nonlinear feedback control design with the Youla-Ku\v{c}era parameterization;
    \item Image classification on the permuted sequential MNIST \cite{le2015simple} and sequential CIFAR-10 problems \cite{chang2019antisymmetricrnn}.
\end{enumerate}
We used the exact same experimental setup as \cite{Revay++2023} for the first two case studies. For the third, we trained controllers for the same linear system and cost function as in \cite{Revay++2023}, but using unrestricted contracting models and analytic-gradient-based reinforcement learning as in \cite{Barbara++2025c}. For the fourth, we followed the setup in \cite{kozachkov2022RNNs} and trained the models as image classifiers by passing in images as sequences of pixels, and using the final model output at the end of each sequence as the predicted class. We trained RENs and R2DNs with similar numbers of learnable parameters. Further setup and training details are provided in Appendix~\ref{app:image-classification} and our code\footnoteref{note1}.

\begin{figure}[!t]
    \centering
    \begin{subfigure}[b]{0.48\linewidth}
        \centering
        \includegraphics[trim={0cm 0.5cm 0cm 0cm},clip,width=\textwidth]{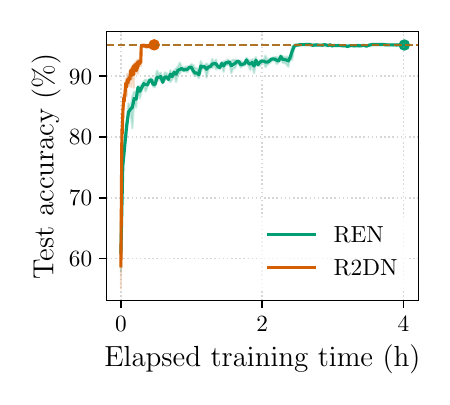}
        \caption{Permuted MNIST.}
        \label{fig:loss-mnist}
    \end{subfigure}
    \begin{subfigure}[b]{0.48\linewidth}
        \centering
        \includegraphics[trim={0cm 0.5cm 0cm 0cm},clip,width=\textwidth]{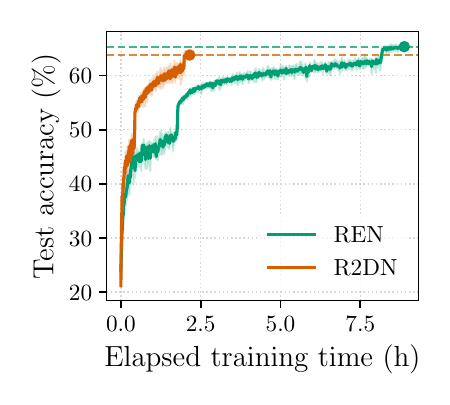}
        \caption{Sequential CIFAR-10.}
        \label{fig:loss-cifar}
    \end{subfigure}
    \caption{Mean test accuracy as a function of training wall time for the sequential image classification tasks. Bands show the range over 3 random seeds. Dashed lines as in Figure~\ref{fig:performance}.}
    \label{fig:performance-cfn}
    \vspace{-3mm}
\end{figure}

Figures~\ref{fig:performance} and \ref{fig:performance-cfn} show performance as a function of training wall time for each experiment. The R2DN models achieve similar losses and classification accuracies to the RENs on most tasks, but are significantly faster to train, even though the model sizes are similar. 
The benefit is most obvious for the system identification, learning-based control, and sequential image classification tasks, since the models were evaluated on long-horizon simulations in each training epoch. For observer design, the models were trained to minimize the one-step-ahead prediction error (see \cite[Sec.~VIII]{Revay++2023}) and so there were fewer model evaluations per epoch.
The boost in computational efficiency is a direct benefit of not having to solve an equilibrium layer every time the model is called, which speeds up both model inference and backpropagation times, since backpropagation through a REN also requires solving an equilibrium layer (Appendix~\ref{app:acyclic-solver}). Indeed, Table~\ref{tab:timing} confirms that R2DNs benefit from consistently faster inference across all tasks, paving the way for their use in applications requiring real-time decision making.

\begin{table}[!b]
    \centering
    \small
    \setlength{\tabcolsep}{4pt}
    \vspace{-2mm}
    \caption{Mean inference time for RENs and R2DNs on each task, averaged over 2000 time steps and across all trained models. For image classification, we report the per-time-step time multiplied by the image sequence length, since a prediction requires reading every pixel in the image. Sequence lengths are 784 pixels for MNIST and 1024 pixels for CIFAR-10.}
    \label{tab:timing}
    \begin{tabular}{l|ccc|cc}
        \toprule
        & Sys. ID & Observer & Control & MNIST & CIFAR-10 \\
        & ($\mu$s) & ($\mu$s) & ($\mu$s) & (ms) & (ms) \\
        \midrule
        REN     & 283 & 375 & 767 & 225 & 157 \\
        R2DN    & 12.6  & 12.5  & 16.8  & 8.25   & 10.8  \\
        \midrule
        Speedup & 22.4$\times$ & 30.0$\times$ & 45.7$\times$ & 27.3$\times$ & 14.5$\times$ \\
        \bottomrule
    \end{tabular}
\end{table}

Note, however, that there appears to be slight performance loss from using R2DN for the system identification and the sequential CIFAR-10 image classification tasks, while R2DN outperforms REN as a PDE observer. We leave a full characterization of how the elements of each model class -- i.e., the equilibrium layer of a REN versus the many forms of 1-Lipschitz nonlinearities available to an R2DN -- capture different types of complex nonlinear behavior to future work.

\section{Conclusions \& Future Work} \label{sec:conc}

This paper has introduced a parameterization of contracting and Lipschitz recurrent models for machine learning and data-driven control. Compared to RENs, R2DNs are more computationally efficient, exhibit similar expressive power, and are compatible with sparse and highly structured DNN representations, without sacrificing stability and robustness. In future work, we will remove the assumption that $D_{22}$ is zero for $\gamma$-Lipschitz R2DNs,  extend the parameterization to $(Q, S, R)$-robust models, and study the scalability and performance of R2DNs in high-dimensional machine learning tasks, particularly those requiring structured network components.

\bibliographystyle{IEEEtran}
\bibliography{references}

\begin{appendices}
\section{Solving Acyclic RENs} \label{app:acyclic-solver}

In Section~\ref{sec:exp-performance}, we compared R2DNs to acyclic RENs \cite[Sec.~III.B]{Revay++2023}. An acyclic REN is defined as a REN \eqref{eqn:ren} with strictly lower-triangular $D_{11}$. The implementations of acyclic RENs in \cite{Revay++2023,Furieri++2024,Barbara++2025c} and similar works all leveraged the lower-triangular structure for efficient inference, but not for backpropagation. We therefore document our procedure for training and inference of acyclic RENs in this appendix.

To simplify notation, re-write \eqref{eqn:implicit-layer} at a given time step as $w = \sigma(Dw + b)$, where $w = w_t$, $D = D_{11}$ is strictly lower-triangular, and $b = C_1 x_t + D_{12} u_t + b_v$. The solution $w^\star$ to the forwards pass (inference) can be computed row-by-row: 
\begin{equation} \label{eqn:acyclic-fwd}
    w_i^\star = \sigma \left( b_i + \sum_{k < i} D_{ik} w_k^\star \right),
\end{equation}
where $i = 1, \ldots, q$ denotes the $i^\mathrm{th}$ element of $w^\star$. We also compute the following intermediate variables for later use:
\begin{equation*}
    v^\star = Dw^\star + b, \quad J = \mathrm{diag}(\sigma'(v_1^\star), \ldots, \sigma'(v_q^\star)),
\end{equation*}
where $J$ is the Clarke generalized Jacobian of the activation function $\sigma$ at $v^\star$.

For the backward pass, we must compute the gradients $\pdv{\mathcal{L}}{D}$, $\pdv{\mathcal{L}}{b}$ of the loss function $\mathcal{L}$ from \eqref{eqn:learning} with respect to the equilibrium layer inputs $D$ and $b$, given knowledge of $\pdv{\mathcal{L}}{w^\star}$. One approach would be to automatically differentiate through the \texttt{for} loop for each $i$ in \eqref{eqn:acyclic-fwd}, as in \cite{Furieri++2024}. Alternatively, one can apply the implicit function theorem to show that
\begin{equation} \label{eqn:ift}
    \pdv{\mathcal{L}}{(\cdot)} = \left(\pdv{\mathcal{L}}{w^\star} \right)^\top \pdv{w^\star}{(\cdot)} = g^\top (I - JD)^{-1} J \left. \pdv{v^\star}{(\cdot)}\right|_{w^\star}
\end{equation}
as per \cite{winston2020Monotone,Revay++2023}, where we let $g = \pdv{\mathcal{L}}{w^\star}$, $(\cdot)$ denotes either $D$ or $b$, and the partial derivative on the far right is evaluated at the solution $w^\star$ computed by the forward pass \eqref{eqn:acyclic-fwd}. This is the approach taken by \cite{Revay++2023,Barbara++2025c}.

By itself, \eqref{eqn:ift} is not necessarily more efficient than automatic differentiation through \eqref{eqn:acyclic-fwd}. However, we can leverage the strictly lower-triangular structure of $D$ to simplify the computation. Defining $\bar{v}^\top := g^\top (I - JD)^{-1}J$, a straightforward calculation shows that
\begin{equation} \label{eqn:implicit-param-grads}
    \pdv{\mathcal{L}}{D} = \bar{v} (w^\star)^\top, \quad \pdv{\mathcal{L}}{b} = \bar{v},
\end{equation}
so we need to compute $\bar{v}$. Noting that $J$ is diagonal, we re-arrange the definition of $\bar{v}$ to get $$\bar{v} = J(g + D^\top \bar{v}),$$ which is another equilibrium (implicit) equation, this time with strictly upper-triangular $D^\top J$, which can be solved row-by-row as
\begin{equation} \label{eqn:vbar-grad}
    \bar{v}_i = \sigma'(v_i^\star) \left(g_i + \sum_{k>i} D_{ki} \bar{v}_k\right).
\end{equation}
Combining \eqref{eqn:implicit-param-grads} and \eqref{eqn:vbar-grad} allows us to compute gradients of the equilibrium layer in a REN with a simple \texttt{for} loop, just like the forward pass. We have implemented this approach in our GitHub repository\footnoteref{note1} released alongside this paper.

\section{Additional Numerical Details} \label{app:experiment-details}

\subsection{Scalability Experiments} \label{app:expressivity-details}

For the numerical results in Section~\ref{sec:exp-scalability}, all models were trained over 1500 epochs using the Adam optimizer \cite{Kingma+Ba2015} with an initial learning rate of $10^{-3}$, which we decreased by a factor of 10 every 500 training epochs to ensure convergence. We uniformly sampled $x \in [-30, 30]$, $u \in [-1, 1]$ in training and test batches of $128 \times 512$ and 2048 samples, respectively. We trained models from each class with $n = 1$ internal state and increasingly large nonlinear components $\phi_\mathrm{eq}, \phi_g$. For RENs, we varied the number of neurons over $q \in [20, 200]$. For R2DNs, we fixed $q = 16$ and designed $\phi_g$ with ten layers each of width $n_h \in [4, 80]$. We trained 5 models for each architecture and size, each with a different random initialization.

\subsection{Sequential Image Classification} \label{app:image-classification}

For the sequential image classification tasks from Section~\ref{sec:exp-performance}, we used a similar experimental setup to \cite{kozachkov2022RNNs}. The neural networks take in images as sequences of pixels, and return a classification from the final output of the sequence. For the permuted MNIST task, pixels were re-ordered with a fixed (random) permutation, held constant across all models and random seeds. Training images for CIFAR-10 were augmented with images produced by random crops and horizontal flips. For each task, we used models of approximately 130,000 learnable parameters: $(n, q) = (96, 152)$ and $(n,q) = (128, 78)$ for RENs on MNIST and CIFAR-10, respectively; and $(n,q) = (96, 76)$ with $\phi_g$ consisting of two layers with width $n_h = 120$ each for R2DNs in both experiments. We used ReLU activations and trained each model to minimize the cross-entropy loss of the final output with the Adam optimizer, using an initial learning rate of $10^{-3}$ and minibatches of 512 images. We trained each model for 150 epochs on MNIST, decreasing the learning rate by a factor of 10 after epochs 90 and 130, and for 600 epochs on CIFAR-10, decreasing it after epochs 120 and 550. We trained 3 models for each architecture on each task, each with a different random initialization.

\end{appendices}

\end{document}